\documentclass{article}
\usepackage{ijcai23}

\usepackage{times}
\usepackage{soul}
\usepackage{url}
\usepackage[hidelinks]{hyperref}
\usepackage[utf8]{inputenc}
\usepackage[small]{caption}
\usepackage{graphicx}
\usepackage{amsmath}
\usepackage{amsthm}
\usepackage{booktabs}
\usepackage{algorithm}
\usepackage{algorithmic}
\usepackage[switch]{lineno}
\usepackage{xcolor}
\usepackage{mathtools}
\usepackage{amssymb}
\usepackage[FIGBOTCAP, TABBOTCAP]{subfigure}
\usepackage{multirow}
\newcommand{\para}[1] {\vspace{6pt}\noindent\textbf{#1}}

\newcommand\paren[1]{\left(#1\right)}
\newcommand\abs[1]{\left\lvert#1\right\rvert}
\newcommand\ceil[1]{\left\lceil#1\right\rceil}

\title{QAQ: Quality Adaptive Quantization for LLM KV Cache}

\author{
    Shichen Dong$^{*1}$, Wen Cheng$^{*1}$, Jiayu Qin$^1$, Wei Wang$^1$
    \affiliations
    $^1$Nanjing University
    \emails
    \{scdong, wcheng, jiayuqin\}@smail.nju.edu.cn, ww@nju.edu.cn
}
\begin{document}

\maketitle
\let\thefootnote\relax\footnotetext{$*$ Co-first author, contributed equally.}
\begin{abstract}
    The emergence of LLMs has ignited a fresh surge of breakthroughs in NLP applications, particularly in domains such as question-answering systems and text generation. 
As the need for longer context grows, a significant bottleneck in model deployment emerges due to the linear expansion of the Key-Value (KV) cache with the context length.
Existing methods primarily rely on various hypotheses, such as sorting the KV cache based on attention scores for replacement or eviction, to compress the KV cache and improve model throughput. 
However, heuristics used by these strategies may wrongly evict essential KV cache, which can significantly degrade model performance.
In this paper, we propose QAQ, a \underline{Q}uality \underline{A}daptive \underline{Q}uantization scheme for the KV cache. 
We theoretically demonstrate that key cache and value cache exhibit distinct sensitivities to quantization, leading to the formulation of separate quantization strategies for their non-uniform quantization. 
Through the integration of dedicated outlier handling, as well as an improved attention-aware approach, QAQ achieves up to $10\times$ the compression ratio of the KV cache size with a neglectable impact on model performance. 
QAQ significantly reduces the practical hurdles of deploying LLMs, opening up new possibilities for longer-context applications.
The code is available at github.com/ClubieDong/KVCacheQuantization.

\end{abstract}

\section{Introduction}
Large Language Models (LLMs) demonstrate state-of-the-art performance on various  Natural Language Processing (NLP) benchmarks \cite{open-llm-leaderboard}. 
These LLMs showcased exceptional potential across a multitude of practical applications, including but not limited to text generation, conversation processing, and knowledge question answering \cite{chang2023survey}. 
However, deploying these models efficiently poses a challenge due to the sequential nature of the generative inference process. 
That is, sequentially processing one token at a time requires accessing all previously generated tokens for computation.
In practical computations, such as GPT series \cite{brown2020language}, LLaMA series \cite{touvron2023llama}, and OPT series \cite{zhang2022opt}, the generative inference of these LLMs typically incorporates the \textit{KV cache} mechanism to improve the efficiency of computation resource utilization.
KV cache stores the computed values of the Key-Value vector from previous attention calculations and reuses them when computing the current token to save extra costs associated with redundant calculations.
While being a widely used optimization method, as the model size and context length continue to increase, the storage overhead of the KV cache itself also grows dramatically, imposing significant pressure on the on-device, especially the high-cost GPU memory. 
\textit{Reducing the memory footprint of the KV cache has become a highly active research topic} \cite{zhu2023survey}.

% \wen{Now revising.}
Currently, there is a substantial body of research addressing the efficient utilization of GPU memory in memory-constrained scenarios.
Offloading is an intuitive solution for handling insufficient GPU memory during model inference. 
Although offloading can alleviate the pressure on GPU memory, implementing offloading specifically for the KV cache is a non-trivial problem, as it is constrained by various factors such as data transmission bandwidth limitations. 
Additionally, approaches like sparse transformers \cite{child2019generating} and multi-query attention \cite{pope2023efficiently} are designed to reduce cache sizes directly; however, applying them directly to optimize the KV cache may result in significant performance degradation \cite{ge2023model}. 
Recently, pioneering studies have emerged, focusing on the direct optimization of the KV cache to minimize its footprint in GPU memory.
However, these methods often rely on attention values to eliminate redundant portions from the KV cache, retaining what is considered essential.
Nevertheless, these approaches may lead to erroneously removing crucial KV cache and significantly degrading the performance of the model \cite{zhang2023h}.
Naturally, it leads to the question: \textit{is there a direct quantization method that can avoid the drawbacks mentioned above, while achieving leading performance?}

In this paper, we propose QAQ, a quality adaptive quantization scheme for KV cache in LLMs.
Quantization is a commonly employed method for compressing model sizes and has been widely utilized for the compression of weights and activations \cite{zhu2023survey}.
However, compressing the KV cache remains a challenging task. 
There are three key insights that inspire QAQ.

\begin{itemize}
    \item First, key cache and value cache exhibit distinct sensitivities to quantization, a proposition validated through theoretical analyses and empirical experiments. 
    This necessitates the development of separate quantization strategies for key cache and value cache.

    \item  Second, the hypothesis of persistence of importance has exceptions. 
    Previous works proposed the hypothesis of the persistence of importance, advocating compression based on importance (attention-aware). 
    We discovered that, despite its validity in the majority of cases, there exist a number of exceptions. 
    This implies the need for careful handling of exceptions when quantizing based on attention.

    \item Third, outliers play a crucial role. 
    While this has been acknowledged in weight and activation quantization, we verified that outliers also exert a significant impact on the KV cache quantization.
    Consequently, a dedicated treatment for quantizing outliers is required.
    
\end{itemize}

With these considerations, QAQ achieves nearly a $10\times$ compression of the KV cache size with minimal impact on model inference performance. 
In comparison to existing attention-based KV cache eviction or replacement approaches, QAQ achieves a further nearly $2\times$ compression.
We make our code publicly available for replication.
\begin{figure}
  \centering
  \includegraphics[width=0.8\linewidth]{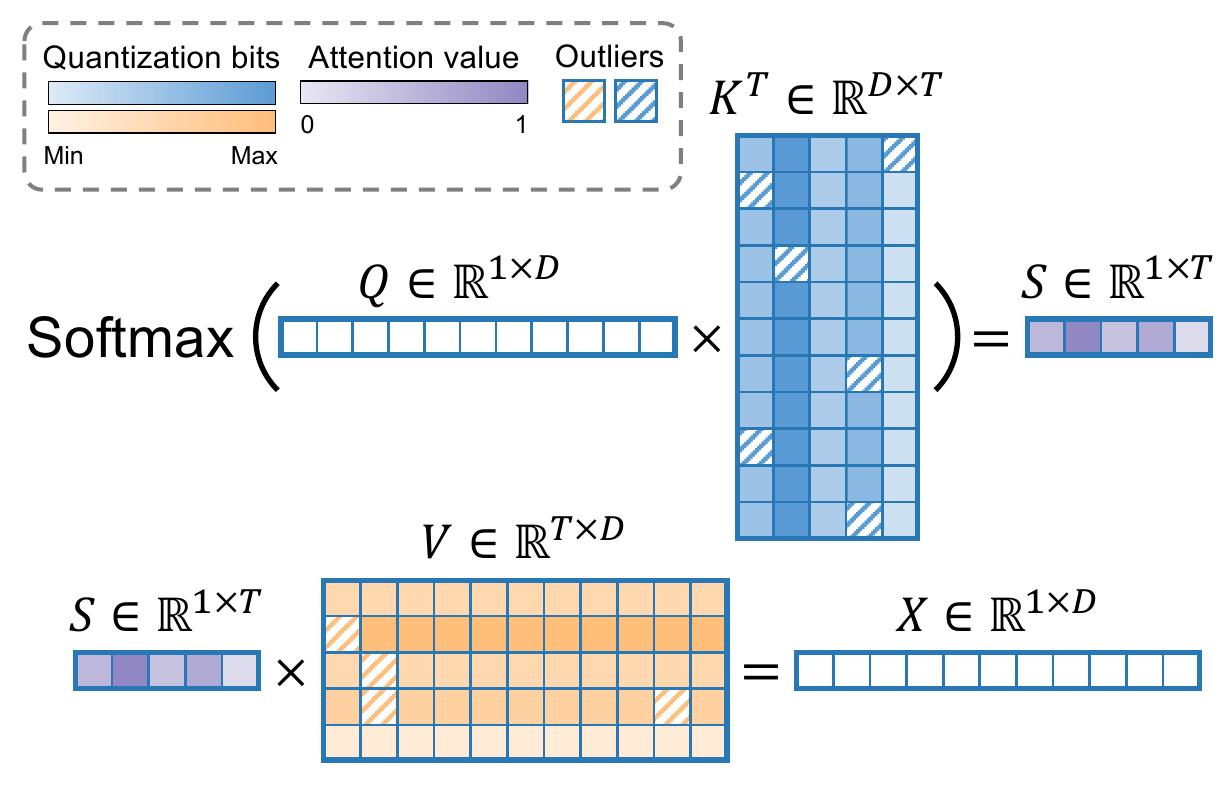}
  \caption{Calculation process of quantized key and value vectors, $\frac{1}{\sqrt{D}}$ is neglected.}
  \label{fig:calculation_process}
\end{figure}
\section{Problem Description and Related Work}
In this section, we first introduce problem profiling, followed by related work.
\subsection{Problem Description}
% \wen{TODO: In process}
The inference process of an auto-regressive generative LLM typically includes two procedures, prompt encoding and token generation. 
In the prompt encoding procedure, for each token generation, the LLM requires contextual information from previous tokens, \emph{i.e.}, key and value vectors (KV vectors). 
KV vectors are stored in the KV cache once they are generated to eliminate redundant computations. 
When a new token is generated in the token generation procedure, its associated KV vectors are appended to the KV cache, which implies that the size of the KV cache linearly increases with the length of the token sequence.
However, the KV cache demonstrates a linear growth relationship with sequence length.
As the model requires longer contexts, the KV cache becomes a substantial performance bottleneck.
Taking OPT-175B as an example, with a total of 96 layers and a hidden size of 12288, its weights occupy 325GB memory, while the KV cache is $3.54\times$ larger, reaching 1152GB under its maximum sequence length.

We formally define the problem of quantization of the KV cache, we focus on the memory footprint of the KV cache in the attention calculation. 
Denote $\textbf{Q} \in \mathbb{R}^{1\times D}$, $\textbf{K} \in \mathbb{R}^{T\times D}$, and $\textbf{V} \in \mathbb{R}^{T\times D}$ as the query tensor, key tensor, and value tensor within each head in every layer, respectively. 
The softmax operation output is denoted by $\textbf{S} \in \mathbb{R}^{1\times T}$, calculated as $\text{Softmax}(\frac{1}{\sqrt{D}} \textbf{Q} \textbf{K}^T)$.
Additionally, $\textbf{X} \in \mathbb{R}^{1\times D}$ represents the product of $\textbf{S}$ and $\textbf{V}$. 
All the notions and their shape are illustrated in Figure \ref{fig:calculation_process}.

For a specific quantization method $C$, we mark the quantized KV cache as $\hat{\textbf{K}}, \hat{\textbf{V}} = f(\textbf{K}, \textbf{V}, C)$, we choose the method that minimizes the loss of accuracy and at the same time using the least memory, as follows.
\begin{small}
\begin{align}
% \scalebox{0.8}{
C^*=\text{argmin}_{C\in \mathcal{C}}\ \text{KVCacheMemory}(C) \quad &\text{s.t.} \nonumber \\
\left\lVert \text{Softmax}\left(\frac{1}{\sqrt{D}}\textbf{Q}\textbf{K}^T\right)-\text{Softmax}\left(\frac{1}{\sqrt{D}} \textbf{Q}\hat{\textbf{K}}^T\right)\right\rVert^2_2 &\leq \Delta_S \nonumber \\
\left\lVert \text{Softmax}\left(\frac{1}{\sqrt{D}} \textbf{Q}\textbf{K}^T\right)\textbf{V} - \text{Softmax}\left(\frac{1}{\sqrt{D}}\textbf{Q}\hat{\textbf{K}}^T\right)\hat{\textbf{V}}\right\rVert^2_2 &\leq \Delta_X\text{,} \nonumber
% }
\end{align}
\end{small}
where $\mathcal{C}$ represents the set of all quantization methods, $\text{KVCacheMemory}(C)$ is the KV cache memory cost, which in our design is calculated by the quantized bit, of method $C$, $\Delta_S$, $\Delta_X$ is the constrained quantization loss for $\textbf{S}$ and $\textbf{X}$, respectively, and are hyper-parameters to control the aim of accuracy. 
\subsection{Related Work}
To alleviate the practical deployment challenges associated with the increasing scale of models, numerous methods have been developed in recent years for model compression. 
The memory footprint of a LLM consists of three components: model weights, activation, and KV cache. 
Early compression techniques primarily targeted the first two components \cite{zhu2023survey}.
Among these, the most representative categories include quantization, pruning, distillation, and low-rank approximation.
%
% \textbf{Quantization.} 
In the field of model compression, quantization is a widely embraced method. 
Quantization involves converting the floating-point representations within the model into discrete forms, such as integers, which can significantly reduce the storage requirement.
Carefully designed quantization methods aim to minimize the accuracy loss to an acceptable range. 
Quantization can be categorized into two main types: quantization-aware training (QAT) and post-training quantization (PTQ). 
QAT is less practical due to the substantial re-training costs, while PTQ without careful design may lead to severe accuracy degradation.
In the early stage of PTQ, certain approaches focus on quantizing only the weight of LLMs. 
OPTQ \cite{frantar2022optq} introduces 3 or 4 bits quantization for weights with improved performance.
%
% LUT-GEMM \cite{park2022nuqmm} optimizes the matrix multiplications for weight-only quantization. 
%
LLM.int8() \cite{dettmers2022llm} exploits the importance of the outliers and employs vector-wise quantization and mixed-precision decomposition for \textit{outliers}. 
AWQ \cite{lin2023awq} finds only $1\%$ of overall weights have a great impact on the performance of the model, it proposes attention-aware quantization based on this insight.
ZeroQuant \cite{yao2022zeroquant} integrates a quantization scheme for both weight and activation in LLMs.
As the model demands higher capabilities for handling extremely long contexts, compressing the KV cache becomes pronounced.
There is a limited body of recent art directly towards compressing the KV cache in LLM to mitigate the bottleneck.
FastGen \cite{ge2023model} develops an adaptive compression method for the KV cache, leveraging the observation that abundant structures exist in attention modules. 
H2O \cite{zhang2023h} exploits the importance of a small set of tokens and proposes an efficient eviction strategy for the KV cache. 
Scissorhands \cite{liu2023scissorhands} validates the persistence of importance hypothesis for the KV cache and reduces the storage buffer.
% \wen{Lack a description of these drawbacks and our advantages.}
However, these methods rely on attention values to eliminate redundant parts in the KV cache, retaining the so-called important portions.
Nevertheless, any misjudgment of importance leading to the loss of crucial cache can significantly degrade the performance of the model.
\begin{figure*}[!ht]
    \centering
    \subfigure[KV cache exhibits different sensitivity to quantization.]{
    \includegraphics[width=0.30\linewidth]{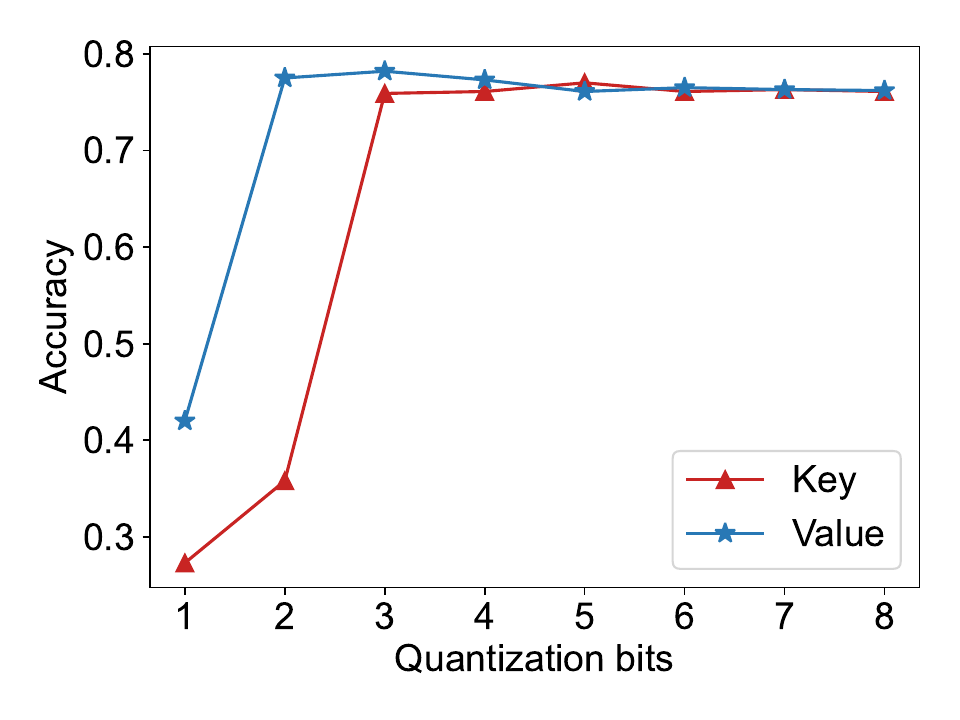}
    \label{fig:kv_cache_is_different}
    }
    \subfigure[Exceptions in attention matrix.]{
    \includegraphics[width=0.30\linewidth]{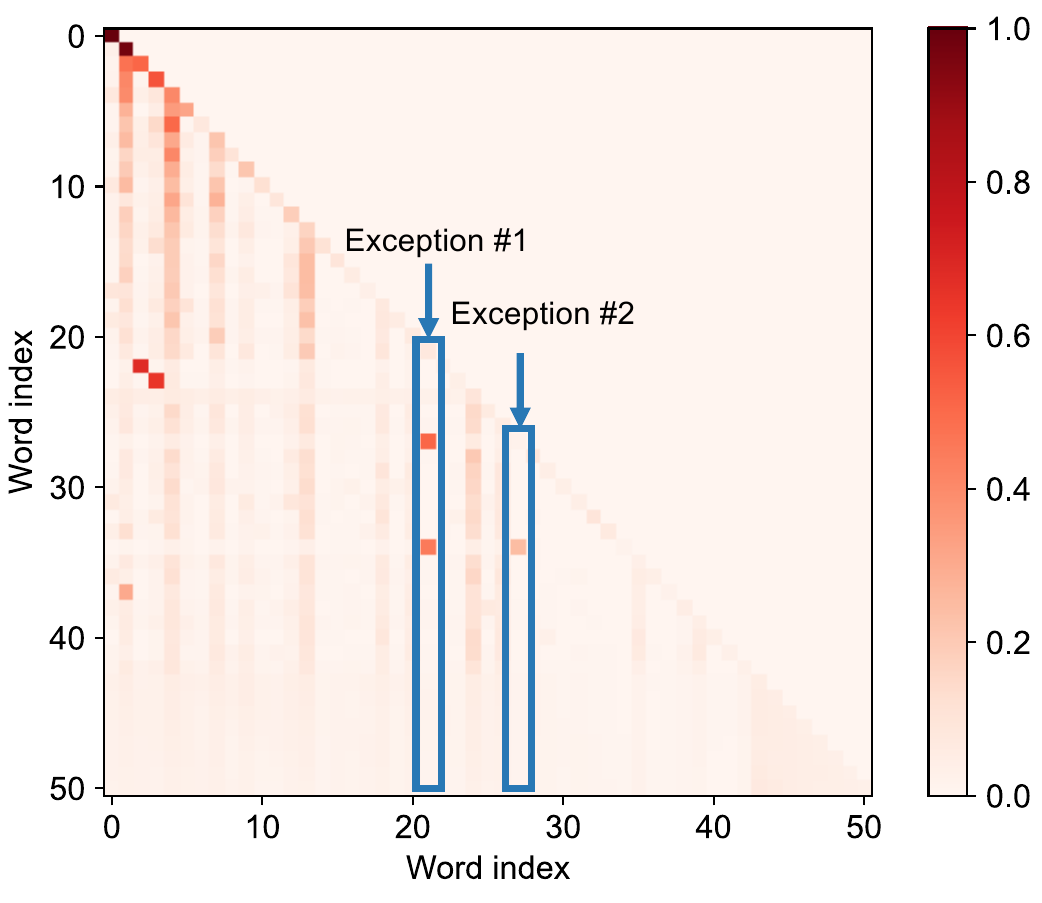}
    \label{fig:exceptions_kv_cache}
    }
    \subfigure[Distribution of KV cache value and outliers.]{
    \includegraphics[width=0.35\linewidth]{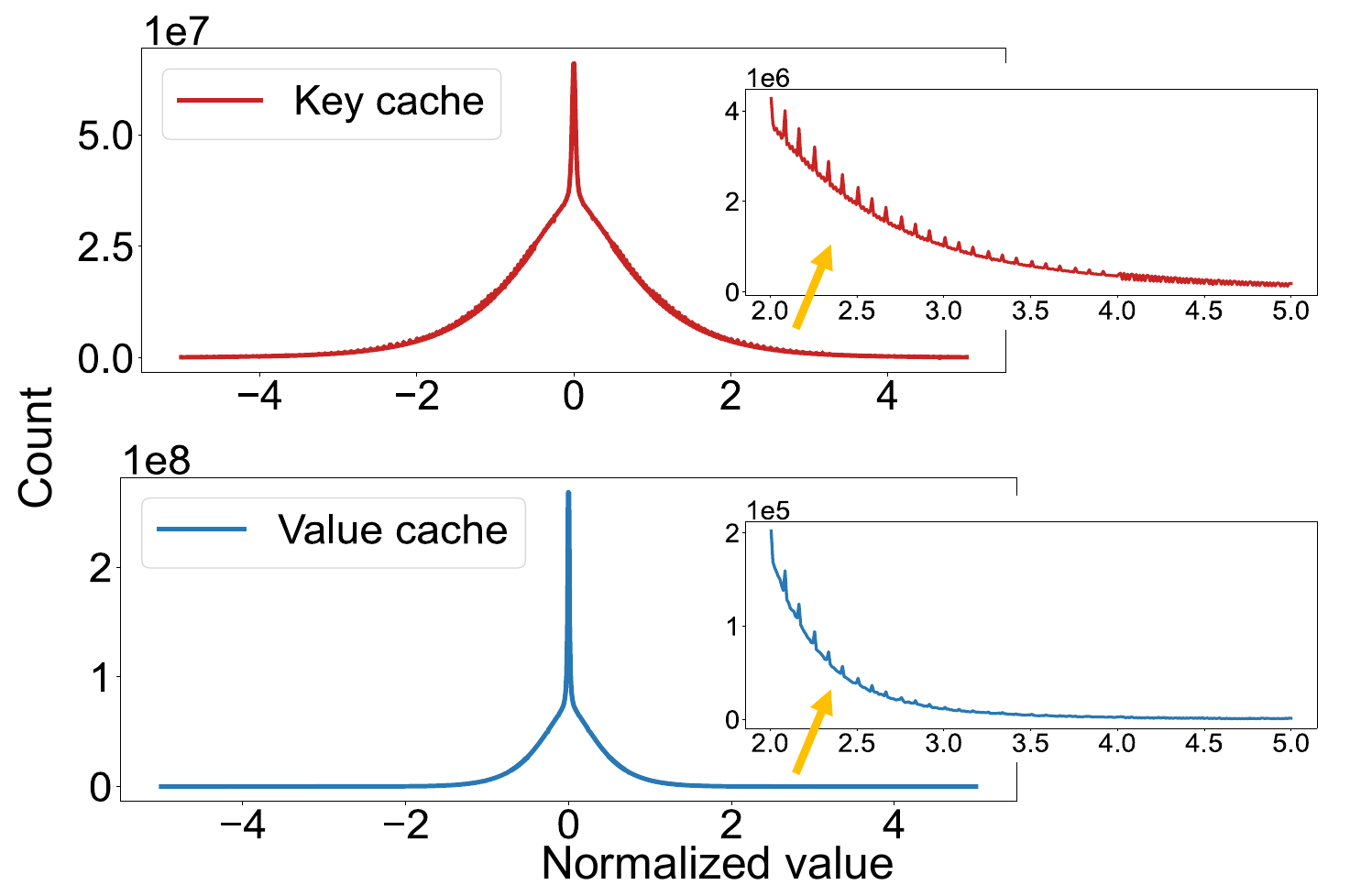}
    \label{fig:kv_cache_distribution_outliers}
    }

    \caption{The experiment demonstration of our key insights.}
\end{figure*}
\section{Insights}
%Insights - Challenges - Solutions
In this section, we will introduce the exposition of the three key insights that inspire our design.
\subsection{Key Cache and Value Cache Exhibit Distinct Sensitivities to Quantization}
Given the distinct roles of key vector and value vector in attention computations, the impact of quantization on the key cache and value cache yields disparate outcomes. 
The theoretical derivation supporting this assertion is as follows.

We first investigate the partial derivative of $X_j$ concerning $V_{ti}$ as follows:
\begin{align*}
\frac{\partial X_j}{\partial V_{ti}} = 
\begin{cases}
0& i\neq j\\
S_t& i=j
\end{cases}
\text{.}
\end{align*}
%
% We begin by theoretically demonstrating this phenomenon.
Similarly, we investigate the partial derivative of $X_j$ with respect to $K_{ti}$ as follows:
\begin{align*}
\frac{\partial X_j}{\partial K_{ti}}
% =&\frac{\partial (\sum_{a=1}^T S_a V_{aj})}{\partial K_{ti}}\\
% =&\sum_{a=1}^T V_{aj}\frac{\partial S_a}{\partial K_{ti}}\\
% % =&S_t\cdot Q_i(V_{tj} - \sum_{a=1}^T V_{aj}S_a)
=&S_t\cdot Q_i(V_{tj} - X_j)
\text{.}
\end{align*}
% \shichen{$\sum_{a=1}^T V_{aj}S_a=X_j$}
% \shichen{Consider remove the calculation process}
%
% The variance  
From the above two equations, it is evident that changes in $K$ have a more pronounced impact on $X$ compared to $V$.
In other words, the key cache is more sensitive to quantization, quantizing key cache results in more severe performance degradation.
To validate our theoretical analysis, we utilize a uniform quantization approach to quantize the key cache and value cache individually in LLaMA~2-7B. Subsequently, we assess the model's accuracy in responding to 1,000 randomly sampled questions from HellaSwag \cite{zellers2019hellaswag}. 
The results are depicted in Figure \ref{fig:kv_cache_is_different}. 
Notably, the key cache and value cache exhibit distinct sensitivities to the same uniform quantization granularity.
Specifically, when uniformly quantizing the value cache to 2 bits only, the model's performance maintains a relatively high accuracy.
However, in the case of only uniformly quantizing the key cache to 2 bits, the model's performance degrades significantly, aligning with our derived conclusions.
This underscores the necessity of employing distinct quantization strategies for the key cache and value cache to acquire the best performance.
To the best of our knowledge, we are the first ever to exploit this observation.
\subsection{Persistence of Importance has Exceptions}
\label{subsec:insight-attention}
The persistence of important tokens, which means the specific tokens that have larger attention value are the only ones significant for now and future steps in the LLM inference process, has been raised in the past works \cite{liu2023scissorhands}. 
Although this finding holds in most cases, there are still some exceptions where tokens that were initially less significant become suddenly crucial in the subsequent process of generation.
%
% \textcolor{red}{It is leading to the low performance of existing attention-based KV cache compression approaches.}

Figure \ref{fig:exceptions_kv_cache} illustrates the attention computations matrix for the $1^{st}$ layer, $8^{th}$ head of LLaMA~2-7B \cite{touvron2023llama}, after the inference process on a randomly selected question from HellaSwag \cite{zellers2019hellaswag}. 
It is evident that the importance of the majority of tokens tends to persist, indicating a relatively stable importance level. 
However, there are exceptions where the importance of a few tokens deviates. 
Specifically, as illustrated by exception \#1 and exception \#2, certain tokens, initially considered less important, undergo a sudden shift in importance during a specific instance of inference. Without additional treatment for these exceptional cases, based on the assumptions of existing methodologies, these exceptional tokens might be evicted or discarded before demonstrating their importance. 
This could result in the loss of information when their importance changes and they are required for computation, subsequently impacting the model's performance.

In our design, we propose the \textit{attention window} as the solution, which stores the maximum value from the previous $n$ attention scores for each token, addressing the exceptional cases mentioned above.
\subsection{Outliers in KV Cache Matter}
\label{subsec:insight-outliers}
The handling of outliers is of paramount importance as they can significantly impact model performance when formulating quantization strategies.
Existing works have demonstrated a profound understanding of outliers in model weights and activations. 
For instance, LLM.int8() \cite{dettmers2022llm} employs vector-wise quantization and mixed-precision decomposition to address outliers in the model's weights.
OWQ \cite{lee2023owq} theoretically analyzes how activation outliers amplify errors in weight quantization.

The outliers in the KV cache also play an important role.
After randomly testing 1,000 questions from HellaSwag using the LLaMA~2-7B model, the normalized numerical distributions of the key cache and value cache are depicted in Figure \ref{fig:kv_cache_distribution_outliers}. 
We have magnified the tail for better visibility. 
Please note that the jagged distribution in the figure is a result of floating-point precision. 
The graph reveals a substantial presence of outliers in both the key cache and value cache. 
Further experiments indicate that neglecting to address these outliers significantly impacts the model's performance.

Our approach involves the implementation of mixed precision, specifically assigning a separate storage precision for outliers during quantization.
This strategy aims to minimize performance degradation attributable to precision loss associated with outlier values.

\section{Quality Adaptive Quantization Method for Key Cache and Value Cache}
\label{sec:method}

In this section, we first derive the formulas for KV cache quantization. Next, we show how this quantification approach is employed in the text generation procedure.

\subsection{Quantitative Formula Derivation}
\label{subsec:method-formula}

\newcommand\meanof[1]{\mu^{\paren{\textbf{#1}}}}
\newcommand\stdof[1]{\sigma^{\paren{\textbf{#1}}}}
\newcommand\varof[1]{\sigma^{2\paren{\textbf{#1}}}}
\newcommand\bitsof[1]{B^{\paren{\textbf{#1}}}}
\newcommand\errorof[1]{\Delta^{\paren{\textbf{#1}}}}
\newcommand\MeanOf[1]{\mathbb{E}\left[#1\right]}
\newcommand\VarOf[1]{\text{Var}\left[#1\right]}

The idea behind the derivation of our formula is as follows: We regard the quantized KV cache ($\hat{\textbf{K}}$ or $\hat{\textbf{V}}$) as a set of independent random variables, with means equivalent to the unquantized KV cache ($\textbf{K}$ or $\textbf{V}$) and customizable standard deviations ($\stdof{K}_t$ or $\stdof{V}_t$) for each token $t$, controlled by the number of quantization bits. Then, we treat the attention value $\hat{\textbf{S}}$ as a function of $\hat{\textbf{K}}$ and the output of the self-attention module $\hat{\textbf{X}}$ as a function of $\hat{\textbf{V}}$, both of which are also considered as random variables. To keep the accuracy optimal, it is desirable that $\hat{\textbf{S}}$ and $\hat{\textbf{X}}$ are close to their unquantized counterparts $\textbf{S}$ and $\textbf{X}$. We achieve this by restricting the standard deviation of $\hat{\textbf{S}}$ or $\hat{\textbf{X}}$ less than a given hyperparameter $\stdof{S}_{\text{max}}$ or $\stdof{X}_{\text{max}}$, which provides an upper bound for $\stdof{K}_t$ or $\stdof{V}_t$, respectively. Finally, we calculate the minimum number of quantization bits of KV cache for each token $t$ that ensures the constraints of $\stdof{K}_t$ or $\stdof{V}_t$ are satisfied.

\para{Value cache quantization.}
We start with the quantization of value cache, which is simpler in comparison. The value at index $d$ in the output of the self-attention module is given by:
$$
\hat{X}_d=\sum_{t=1}^T S_t\cdot\hat{V}_{td}\text{,}
$$
where the standard deviation of $\hat{V}_{td}$ is $\stdof{V}_t$ for each $t=1,\ldots,T$ and $d=1,\ldots,D$. Therefore, the variance of $\hat{X}_d$ can be expressed as:
$$
\varof{X}_d=\sum_{t=1}^T S_t^2\cdot\varof{V}_t\text{,}
$$
which implies that the error of $X_d$ accumulates from the error of each token. Ideally, the error contribution from each token, $S_t^2\varof{V}_t$, should be uniform across all tokens. Furthermore, to ensure that $\stdof{X}_d$ remains below the given upper bound $\stdof{X}_{\text{max}}$, the following constraint must be satisfied:
$$
\stdof{V}_t\le\frac{1}{\sqrt{T}}\cdot\frac{\stdof{X}_\text{max}}{\abs{S_t}}\text{.}
$$
This formula suggests that the error of value cache at each token should be inversely proportional to its corresponding attention value.

\para{Key cache quantization.}
Due to the involvement of the Softmax function, the quantization of key cache is inherently more complex. The attention score of token $t$ is given by:
$$
\hat{S}_t=\frac{e^{\hat{A}_t}}{\sum_{i=1}^T e^{\hat{A}_i}}\text{, where }\hat{A}_t=\sum_{d=1}^D Q_d\cdot\hat{K}_{td}\text{.}
$$
Given that $\hat{A}_t$ is the sum of a collection of independent random variables, it follows the normal distribution $N(\meanof{A}_t,\stdof{A}_t)$ according to the central limit theorem, where:
$$
\meanof{A}_t=\sum_{d=1}^D Q_d\cdot K_{td}\text{ and }
\varof{A}_t=\sum_{d=1}^D Q_d^2\cdot\varof{K}_t\text{.}
$$
By definition, $e^{\hat{A}_t}$ follows the log-normal distribution $\text{LogNormal}(\meanof{A}_t,\stdof{A}_t)$ with the mean and variance being:
\begin{align*}
\MeanOf{e^{\hat{A}_t}}&=e^{\meanof{A}_t+\varof{A}_t/2}\text{,}\\
\VarOf{e^{\hat{A}_t}}&=e^{2\meanof{A}_t+\varof{A}_t}\paren{e^{\varof{A}_t}-1}\text{.}
\end{align*}
Similarly, we expect the uniformity in error contribution from each token, implying that $e^{\hat{A}_t}$ should be identically distributed for all $t=1,\ldots,T$. $\hat{S}_t$ is the ratio distribution between $e^{\hat{A}_t}$ at token $t$ and the sum of $e^{\hat{A}_t}$ over all tokens. Applying a second-order Taylor expansion to the ratio~\cite{seltman2012approximations}, the variance of $\hat{S}_t$ is approximately:
\begin{align*}
\varof{S}_t\approx&
\frac{1}{T^2}\cdot\paren{1-\frac{1}{T}}\cdot\frac{\paren{\MeanOf{e^{\hat{A}_t}}}^2}{\VarOf{e^{\hat{A}_t}}}\\
=&\frac{1}{T^2}\cdot\paren{1-\frac{1}{T}}\cdot\paren{e^{\sum_{d=1}^D Q_d^2\cdot\varof{K}_t}-1}.
\end{align*}
To ensure that $\varof{S}_t$ does not exceed the given upper bound $\varof{S}_\text{max}$, we have:
$$
\varof{K}_t\le\frac{1}{\sum_{d=1}^D Q_d^2}\cdot\log\paren{\frac{T^3}{T-1}\cdot\varof{S}_\text{max}+1}\text{.}
$$
This inequality suggests that the upper bound of $\stdof{K}_t$ is dependent on the squared norm of query tensors (\emph{i.e.}, $\sum_{d=1}^D Q_d^2$), which varies with each inference. To address this variability, we precompute the distribution of the squared norm of query tensors and use the upper 10\% quantile of this distribution in the formula. In this way, the inequality holds in 90\% of cases.

\para{Determining quantization bits.}
In the final step, we determine the quantization bits of KV cache for each token $t$ ($\bitsof{K}_t$ and $\bitsof{V}_t$) based on the upper bound standard deviations of quantized KV cache ($\stdof{K}_t$ and $\stdof{V}_t$) calculated above. Owing to the symmetry between the key cache and value cache in this step, we only demonstrate the derivation using key cache $\textbf{K}$.

To quantize key cache into $\bitsof{K}_t$ bits, we uniformly split the range $\left[K_t^\text{min},K_t^\text{max}\right]$ into $2^{\bitsof{K}_t}$ segments, where $K_t^\text{min}$ and $K_t^\text{max}$ represents the min and max value of key cache at token $t$. We then round values in each segments to the corresponding midpoint. Using this quantization method, the quantization errors of key cache are uniformly distributed in $\left[-\errorof{K}_t,\errorof{K}_t\right]$, where $\errorof{K}_t$ satisfies:
$$
2\cdot\errorof{K}_t\cdot2^{\bitsof{K}_t}=K_t^\text{max}-K_t^\text{min}\text{.}
$$
Given that the standard deviation of the above uniform distribution $\stdof{K}_t$ is $\frac{1}{\sqrt{3}}\errorof{K}_t$, we get:
$$
\bitsof{K}_t=\ceil{\log_2\paren{\frac{K_t^\text{max}-K_t^\text{min}}{2\sqrt{3}\cdot\stdof{K}_t}}}\text{.}
$$

\subsection{Methods}
\label{subsec:method-system}

\para{Attention score prediction.}
Our quantization method necessitates the attention scores that quantify the degree to which future tokens attend to preceding tokens. However, these attention scores are not available at the time of quantization. To address this limitation, we invoke the \emph{persistence of importance} theory, which posits that the attention scores (\emph{i.e.}, importance) of each token remain relatively constant (\emph{i.e.}, persistence) throughout the process of token generation. Consequently, we can approximate future attention scores by using the current ones, and use it in the quantization formulas.

Nevertheless, as stated in Section~\ref{subsec:insight-attention}, certain attention scores exhibit abrupt increments, which poses challenges to the quantization method, since quantization is irreversible, we can only quantize caches from higher to lower bits and not vice versa. To mitigate this issue, we propose \emph{attention window}, a method that keeps track of the attention scores of each token and predicts the future attention scores as the maximum value within a window of the preceding $n$ scores. This strategy ensures that aggressive quantization to lower bits is undertaken only after a sequence of consistently low attention scores has been observed, thereby being confident that future scores will remain low.

\para{Outliers.}
Outliers of KV cache have a significant impact on the model's performance, as highlighted in Section~\ref{subsec:insight-outliers}. We define outliers as the values in the KV cache that exceed the $\alpha\%$ quantile at both the maximum and minimum ends, where $\alpha$ is a hyperparameter that controls the proportion of values deemed as outliers. To mitigate the impact of outliers, we introduce a mixed-precision quantization approach. Specifically, we keep outliers unquantized and store them in a sparse matrix at full precision. 

Compared to quantizing KV cache uniformly, the benefits of this method are twofold: 1) the important outliers themselves are stored accurately without quantization error; 2) the quantization of the remaining values in KV cache can be more granular because the distribution range is significantly reduced without outliers. Our experiments also demonstrate that this method effectively avoids the performance degradation caused by quantization.

\para{Integration in text generation process.}
Current autoregressive LLMs generate text token-by-token. Within each inference iteration with our quantization method integrated, the model takes in a new token, combined with the quantized KV cache of previous tokens, and outputs the KV cache of the new token. We copy the unquantized new KV cache into CPU memory for future use, and calculate the quantization bits for all existing tokens using the previously derived formula. For tokens whose newly-calculated quantization bits are lower than the current bits, we further quantize them to the lower bits. For those otherwise, we take the unquantized KV cache from CPU memory and re-quantize it to the required bits. Although this method introduces additional transfers between CPU and GPU memory, our attention window technique can effectively reduce this overhead though cautious quantization strategy.

\begin{table*}[!ht]
\begin{center}
\scalebox{0.75}{
    \begin{tabular}{c|c|c|c}
% \caption{The performance comparison between QAQ with SOTA methods.}
\hline
Methods                          & Backbone model               & Task                & Compression ratio with less than $1\%$ acc. drop \\ \hline
\multirow{6}{*}{Scissorhands\cite{liu2023scissorhands}} & \multirow{3}{*}{OPT-6B}      & HellaSwag-Five shot & 3                                  \\
                              &                              & PIQA-Five shot               & 5                                  \\
                              &                              & MathQA-Five shot              & 5                                  \\ \cline{2-4} 
                              & \multirow{3}{*}{OPT-13B}     & HellaSwag-Five shot           & 5                                  \\
                              &                              & PIQA-Five shot                & 5                                  \\
                              &                              & MathQA-Five shot              & 5                                  \\ \hline
\multirow{2}{*}{H2O\cite{zhang2023h}}          & \multirow{2}{*}{OPT-30B}     & PIQA                & 5                                  \\
                              &                              & MathQA              & 5                                  \\ \hline
\multirow{6}{*}{QAQ}          & \multirow{3}{*}{LLaMA 2-7B}  & HellaSwag-Zero shot & 7.477                              \\
                              &                              & PIQA-Zero shot      & 7.477                                 \\
                              &                              & MathQA-Zero shot    & 6.036                                  \\ \cline{2-4} 
                              & \multirow{3}{*}{LLaMA 2-13B} & HellaSwag-Zero shot & 8.394                                 \\
                              &                              & PIQA-Zero shot      & 9.024                                 \\
                              &                              & MathQA-Zero shot    & 6.056                                 \\ \hline
\end{tabular}
}
\caption{The experiment performance of QAQ vs. SOTA methods.}
\label{table:qaq_vs_sota_methods}
\end{center}
\end{table*}

\begin{table*}[!ht]
\begin{center}
    \scalebox{0.8}{

\begin{tabular}{cccccc}
\hline
\multicolumn{1}{c|}{Model}                            & \multicolumn{1}{c|}{Task}          & \multicolumn{1}{c|}{w/o outliers acc.} & \multicolumn{1}{c|}{w/o outliers compression ration} & \multicolumn{1}{c|}{w 1\% outliers acc.} & w 1\% outliers compression ration \\ \hline
\multicolumn{1}{c|}{\multirow{3}{*}{LLaMA~2-7B}}  & \multicolumn{1}{c|}{HellaSwag} & \multicolumn{1}{c|}{0.572}            & \multicolumn{1}{c|}{7.981}                           & \multicolumn{1}{c|}{0.722}              & 7.629                             \\
\multicolumn{1}{c|}{}                            & \multicolumn{1}{c|}{PIQA}      & \multicolumn{1}{c|}{0.707}            & \multicolumn{1}{c|}{7.695}                           & \multicolumn{1}{c|}{0.763}              & 7.333                             \\
\multicolumn{1}{c|}{}                            & \multicolumn{1}{c|}{MathQA}    & \multicolumn{1}{c|}{0.250}            & \multicolumn{1}{c|}{7.969}                           & \multicolumn{1}{c|}{0.279}              & 7.497                             \\ \hline
\multicolumn{1}{c|}{\multirow{3}{*}{LLaMA~2-13B}} & \multicolumn{1}{c|}{HellaSwag} & \multicolumn{1}{c|}{0.660}            & \multicolumn{1}{c|}{7.938}                           & \multicolumn{1}{c|}{0.766}              & 7.583                             \\ %\cline{2-6} 
\multicolumn{1}{c|}{}                            & \multicolumn{1}{c|}{PIQA}      & \multicolumn{1}{c|}{0.737}            & \multicolumn{1}{c|}{7.614}                           & \multicolumn{1}{c|}{0.789}              & 7.223                             \\ %\cline{2-6} 
\multicolumn{1}{c|}{}                            & \multicolumn{1}{c|}{MathQA}    & \multicolumn{1}{c|}{0.257}            & \multicolumn{1}{c|}{7.928}                           & \multicolumn{1}{c|}{0.287}              & 7.453                             \\ \hline
\end{tabular}
    }
\end{center}
\caption{The ablation result of outliers in QAQ.}
\label{table:outliers}
\end{table*}

\section{Evaluation}
In this section, we first introduce the experiment setting then we present the results that show QAQ archives up to near $10\times$ compression of KV cache memory footprint with no compromise in accuracy.

\subsection{Experiment Settings}
Our experiments are based on a representative LLM model family, LLaMA~2 with model sizes ranging from 7 billion and 13 billion. 
The outliers ratio is set as $1\%$, and the size of attention window is set as $5$.
We compare the accuracy of QAQ-LLaMA against the original LLaMA on a number of downstream tasks: HellaSwag \cite{zellers2019hellaswag}, MathQA \cite{amini2019mathqa}, PIQA \cite{bisk2020piqa}. 
The evaluation uses a similar architecture as lm-eval-harness \cite{gao2021framework} with zero shot in every task, all experiments are conducted on a server with 8 NVIDIA V100 32GB GPUs.

\subsection{Compression Ratio vs. Accuracy}
\begin{figure}[!ht]
    \centering
    \subfigure[LLaMA 2-7B Zero shot.]{
    \includegraphics[width=0.47\linewidth]{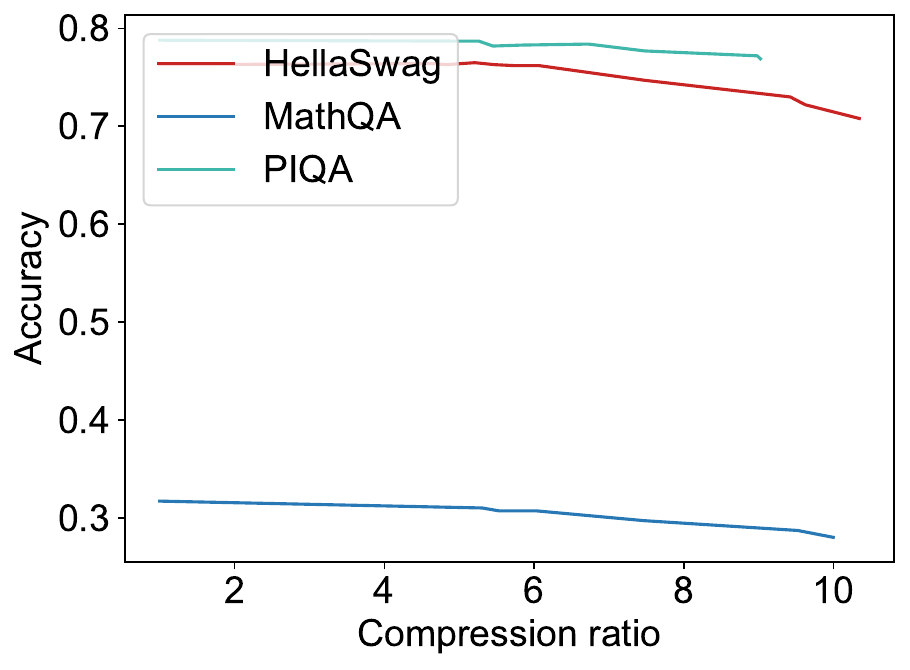}
    \label{fig:downstream_7b}
    }
    \subfigure[LLaMA 2-13B Zero shot.]{
    \includegraphics[width=0.47\linewidth]{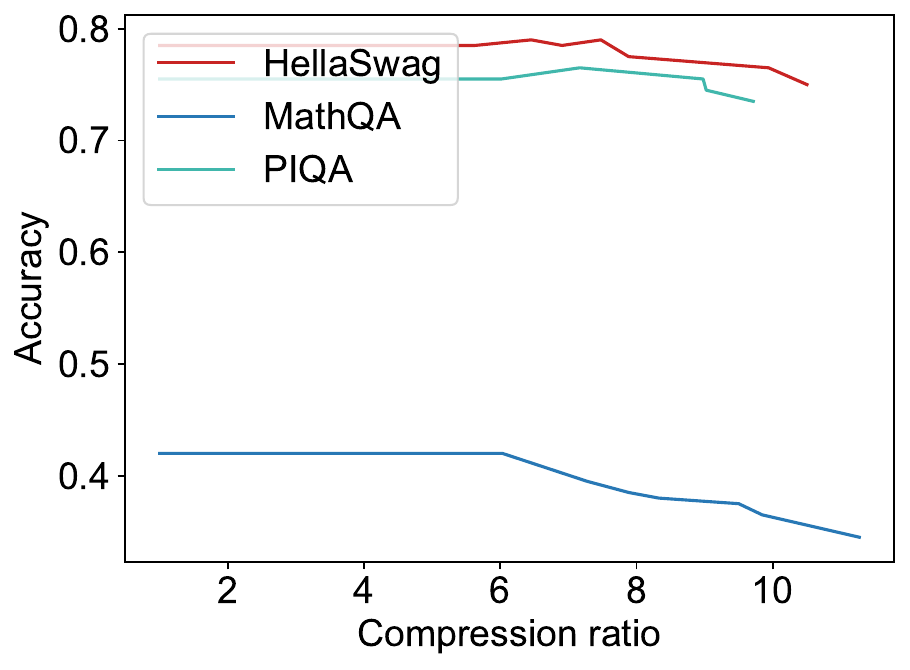}
    \label{fig:downstream_13b}
    }
    \caption{Experiment performance of QAQ.}
    \label{fig:downstream_task}
\end{figure}

We present the experimental evaluation results depicting the variations in accuracy and compression ratio for QAQ in Figure \ref{fig:downstream_task}. 
The compression ratio is defined as the average ratio of the quantized KV cache size to the original KV cache size, where 1$\times$ denotes the original uncompressed LLaMA model.
For both 7B and 13B model specifications, a relatively smooth curve is observed in the graph across four commonly employed downstream tasks. 
Even with compression ratios reaching up to $8\times$, there is minimal impact on the model's performance. 
Remarkably, at compression ratios as high as $10\times$, the overall performance of the model remains exceptionally high. 
Notably, in tasks such as PIQA, the $10\times$ compressed model's performance approaches that of the uncompressed model, indicating the robust capability of QAQ to significantly compress the KV cache size without compromising performance.

\subsection{Comparison}
\begin{figure*}[!ht]
    \centering
    \subfigure[HellaSwag LLaMA 2-7B.]{
    \includegraphics[width=0.3\linewidth]{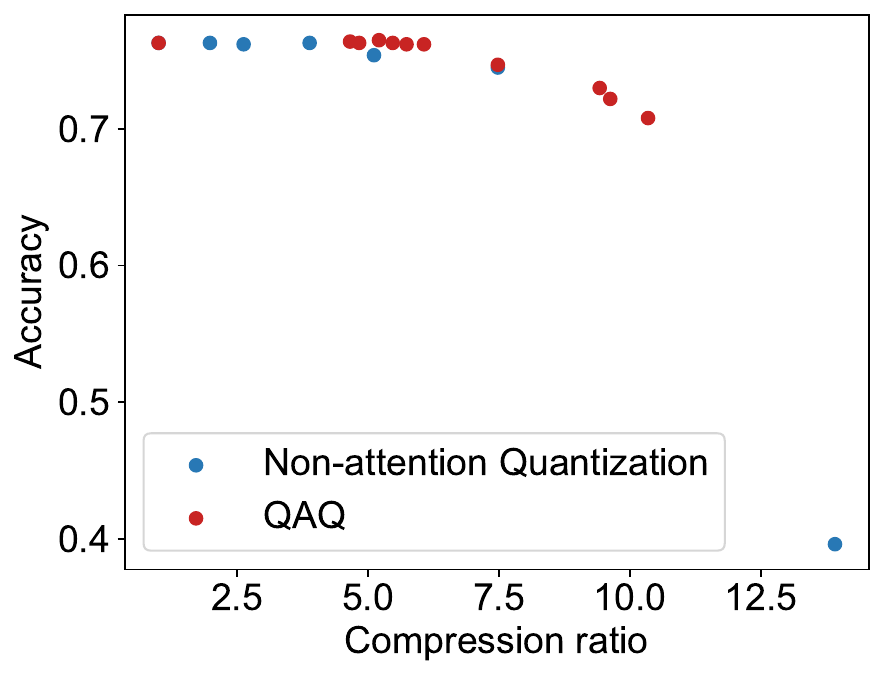}
    % \label{fig:kv_cache_is_different}
    }
    \subfigure[MathQA LLaMA 2-7B.]{
    \includegraphics[width=0.3\linewidth]{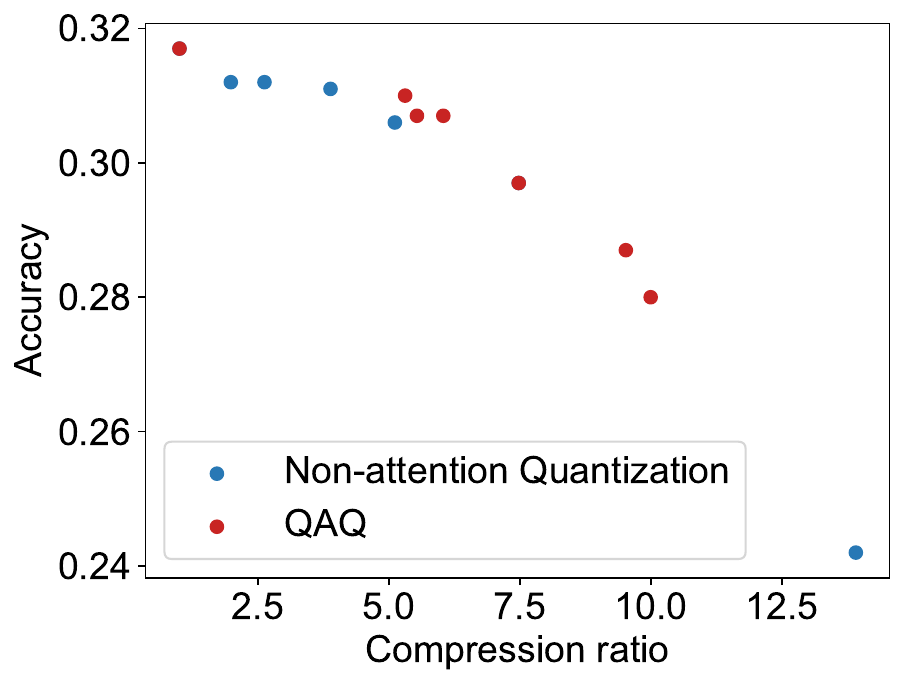}
    % \label{fig:kv_cache_distribution_outliers}
    }
    \subfigure[PIQA LLaMA 2-7B.]{
    \includegraphics[width=0.3\linewidth]{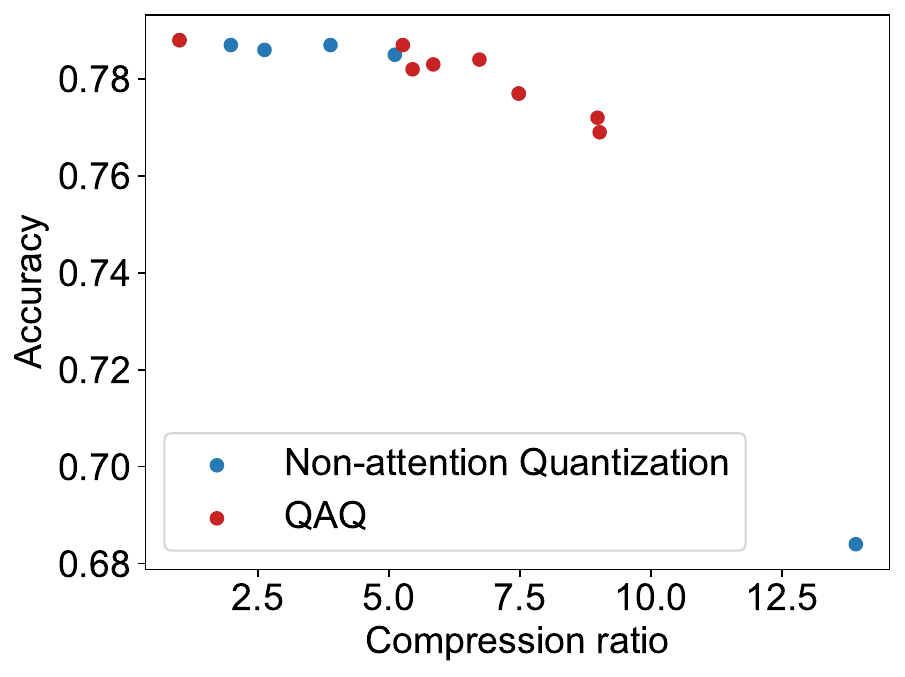}
    % \label{fig:exceptions_kv_cache}
    }
    \subfigure[HellaSwag LLaMA 2-13B.]{
    \includegraphics[width=0.3\linewidth]{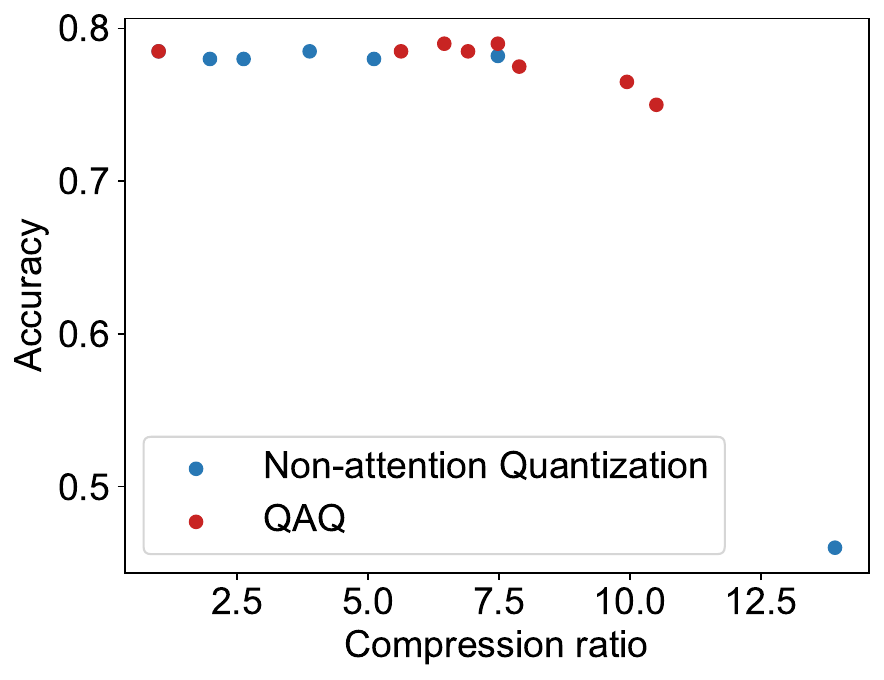}
    % \label{fig:kv_cache_distribution_outliers}
    }
    \subfigure[MathQA LLaMA 2-13B.]{
    \includegraphics[width=0.3\linewidth]{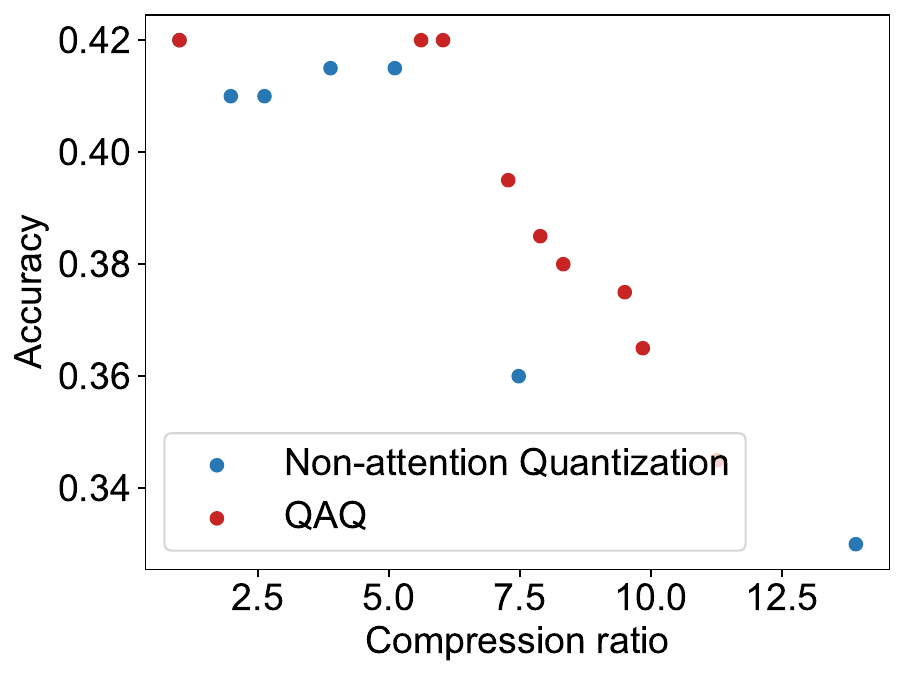}
    % \label{fig:exceptions_kv_cache}
    }    
    \subfigure[PIQA LLaMA 2-13B.]{
    \includegraphics[width=0.3\linewidth]{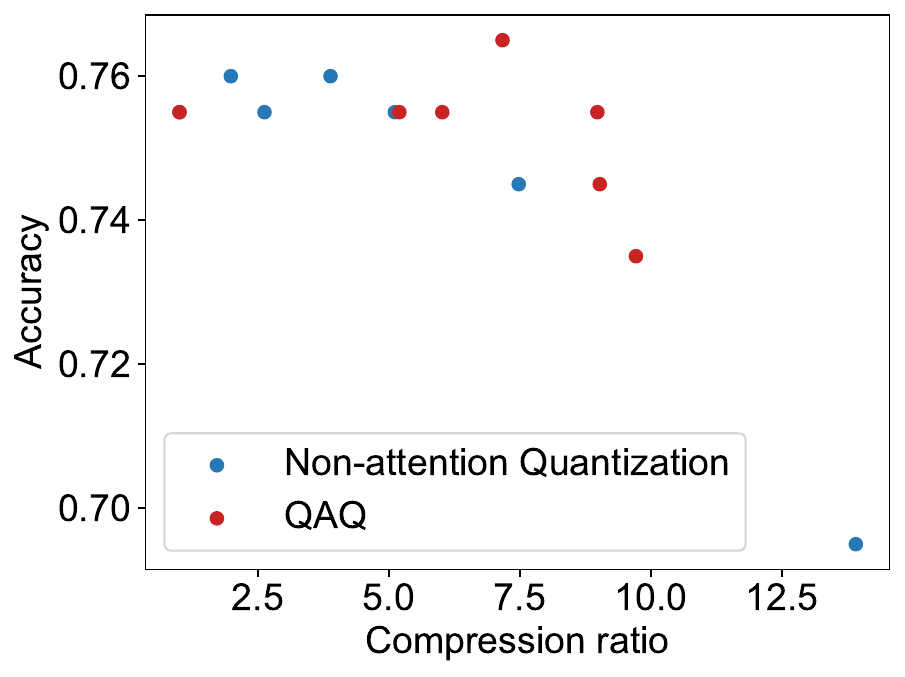}
    % \label{fig:kv_cache_distribution_outliers}    
    }
    \caption{Experiment result of QAQ v.s. non-attention quantization.}
    \label{fig:qaq_vs_uniform}
\end{figure*}

% \wen{Please help examine the name of "uniform", and do we need to move this part to ablation?}
We conducted a comparative analysis with existing state-of-the-art compression methodologies. 
In the realm of model quantization, the widespread practice of uniformly quantizing parameters has exhibited notable compression efficacy \cite{frantar2022gptq}.
Considering the pivotal role played by outliers during quantization, we retained outliers within the framework of uniform quantization. 
The performance evaluation of QAQ is juxtaposed with that of the uniform quantization approach, as illustrated in Figure \ref{fig:qaq_vs_uniform}.
The compression ratio in the scatter plot is determined by the ratio of the compressed KV cache size to the original KV cache size. 
%
% The $1\times$ denotes the original uncompressed LLaMA~2 model. 
%
Closer proximity of data points to the upper-right corner of the figure indicates higher compression ratios and better performance guarantees, signifying superior overall performance.
It is evident that the envelope formed by the QAQ, which is colored in red, data points encompasses the blue data points across all tasks in both models. 
This implies that the LLaMA models quantized using QAQ, whether in the 7B or 13B parameter specifications, exhibit significantly superior performance in the four downstream tasks compared to uniform quantization.
This underscores the exceptional efficacy of the QAQ quantization strategy.
For the existing SOTA techniques in compressing KV cache, we present a comparative analysis of QAQ in Table \ref{table:qaq_vs_sota_methods}. 
It is evident that QAQ achieves a notable $1.6-1.8\times$ improvement in lossless compression ratio across multiple downstream tasks when compared to the current SOTA methods. 
%
% Moreover, QAQ mitigates the potential risk of performance degradation associated with the loss of critical caches. 
%
The results substantiate the outstanding performance of QAQ.

\subsection{Ablations}
To verify the crucial role of outliers in quantization, we conducted ablation experiments while keeping the remaining experimental conditions unchanged.

\para{Outliers.}
To evaluate the outliers' role in the quantization method, we conduct the ablation experiment on the outliers.
% , specifically exploring the impact of including or excluding outliers.
%
The experiments were conducted on the same three downstream tasks, distinguishing between two groups: one where outliers were treated separately and another where no special treatment was applied. 
The presence of outliers was determined based on the numerical distribution.
The experimental results were averaged over 10 trials and are presented in Table \ref{table:outliers}. 
% The experimental results are presented in Table \ref{table:outliers}. 
%
The experimental results indicate that outliers have a substantial impact on KV cache quantization. 
In cases where outliers are not handled separately, the model's performance on downstream tasks experiences a significant decline of $12\% - 26\%$. 
Conversely, treating outliers individually incurs only a $4\%$ additional overhead on the compression ratio.
This demonstrates the efficient and accurate handling of outliers by QAQ.
%
% The result underscores the correctness of QAQ in handling outliers.
%
%

\para{Attention window size.}
To verify the importance of handling exceptional cases in quantization, we conducted ablation experiments to investigate the impact of treating such cases differently. 
The experiments are conducted on the same three downstream tasks, involving two groups: one with a specified window size and the other without, where the absence of a specific setting implies a default window size of $1$.
The results were averaged over 10 trials and are presented in Table \ref{table:last_n_attention}.
The experimental results are presented in Table 3. 
The findings indicate that, across the three downstream tasks, handling exceptional cases results in an improvement of approximately $2\% - 4\%$ in performance. 
This further advances the performance of QAQ.
\begin{table}[!h]
\scalebox{0.7}{
% \usepackage{multirow}
\begin{tabular}{c|c|ccc}
\hline
Model                       & Task                       & Attention window size & Acc.   & Compression ratio \\ \hline
\multirow{6}{*}{LLaMA~2-7B}  & \multirow{2}{*}{HellaSwag} & 1                  & 0.722 & 7.628             \\ \cline{3-5} 
                            &                            & 5                  & 0.730 & 7.377             \\ \cline{2-5} 
                            & \multirow{2}{*}{PIQA}      & 1                  & 0.755 & 7.820             \\ \cline{3-5} 
                            &                            & 5                  & 0.778 & 6.797             \\ \cline{2-5} 
                            & \multirow{2}{*}{MathQA}    & 1                  & 0.276 & 7.633             \\ \cline{3-5} 
                            &                            & 5                  & 0.284 & 7.331             \\ \hline
\multirow{6}{*}{LLaMA~2-13B} & \multirow{2}{*}{HellaSwag} & 1                  & 0.766 & 7.583             \\ \cline{3-5} 
                            &                            & 5                  & 0.772 & 7.340             \\ \cline{2-5} 
                            & \multirow{2}{*}{PIQA}      & 1                  & 0.788 & 7.692             \\ \cline{3-5} 
                            &                            & 5                  & 0.794 & 6.802             \\ \cline{2-5} 
                            & \multirow{2}{*}{MathQA}    & 1                  & 0.283 & 7.588             \\ \cline{3-5} 
                            &                            & 5                  & 0.295 & 7.321             \\ \hline
\end{tabular}
}
\caption{The ablation result of attention window in QAQ.}
\label{table:last_n_attention}
\end{table}
\section{Conclusion}
Inspired by three key insights, we propose a quality adaptive quantization scheme, QAQ, for the KV cache to reduce its memory footprint. 
Our method demonstrates memory reduction of $10\times$ in the KV cache with neglectable loss of accuracy. 
%
% Future exploration including, QAQ is compatible with other model compression techniques, such as pruning and distillation.
%
The superior performance achieved by QAQ allows the model to accommodate longer contextual inputs, creating new possibilities for a broader range of applications.

\appendix

% \section*{Ethical Statement}

% There are no ethical issues.

% \section*{Acknowledgments}

% The preparation of these instructions and the \LaTeX{} and Bib\TeX{}
% files that implement them was supported by Schlumberger Palo Alto
% Research, AT\&T Bell Laboratories, and Morgan Kaufmann Publishers.
% Preparation of the Microsoft Word file was supported by IJCAI.  An
% early version of this document was created by Shirley Jowell and Peter
% F. Patel-Schneider.  It was subsequently modified by Jennifer
% Ballentine, Thomas Dean, Bernhard Nebel, Daniel Pagenstecher,
% Kurt Steinkraus, Toby Walsh, Carles Sierra, Marc Pujol-Gonzalez,
% Francisco Cruz-Mencia and Edith Elkind.

\newpage
\bibliographystyle{named}
\bibliography{ijcai23}

\end{document}